\documentclass{svproc}
\usepackage{url}
\usepackage{comment}

\usepackage[numbers]{natbib} 
\usepackage{url} 
\usepackage{amsmath} 
\usepackage[a4paper, 
            left=2.5cm, 
            right=2.5cm, 
            top=2.5cm, 
            bottom=2cm, 
            footskip=1cm]{geometry}
\usepackage{graphicx}
\usepackage{subcaption}
\usepackage{wrapfig}
\usepackage{tcolorbox}     
\usepackage[font=small]{caption}
\usepackage{subcaption}
\usepackage{tabularx}
\usepackage{amsmath}

\begin{document}

\mainmatter              
%
\title{F-ANcGAN: An Attention-Enhanced Cycle Consistent Generative Adversarial Architecture for Synthetic Image Generation of Nanoparticles }
\titlerunning{Nanoparticle Image Generation}  
\author{Varun Ajith\inst{1} \and Anindya Pal\inst{1},
Saumik Bhattacharya \inst{2} \and Sayantari Ghosh \inst{1}}
\authorrunning{Varun et al.} 
%
%
\institute{National Institute of technology Durgapur, Department of Physics\\
\email{sghosh.phy@nitdgp.ac},
\and
Indian Institute of Technology Kharagpur}

\maketitle              

\begin{abstract}
Nanomaterial research is becoming a vital area for energy, medicine, and materials science, and accurate analysis of the nanoparticle topology is essential to determine their properties. Unfortunately, the lack of high-quality annotated datasets drastically hinders the creation of strong segmentation models for nanoscale imaging. To alleviate this problem, we introduce F-ANcGAN, an attention-enhanced cycle consistent generative adversarial system that can be trained using a limited number of data samples and generates realistic scanning electron microscopy (SEM) images directly from segmentation maps. Our model uses a Style U-Net generator and a U-Net segmentation network equipped with self-attention to capture structural relationships and applies augmentation methods to increase the variety of the dataset. The architecture reached a raw FID score of 17.65 for $\mathrm{TiO_2}$ dataset generation, with a further reduction in FID score to nearly 10.39 by using efficient post-processing techniques. By facilitating scalable high-fidelity synthetic dataset generation, our approach can improve the effectiveness of downstream segmentation task training, overcoming severe data shortage issues in nanoparticle analysis, thus extending its applications to resource-limited fields. 
\end{abstract}
\section{Introduction}
%
Generative adversarial networks (GANs) \cite{goodfellow2014generative} have revolutionized the process of unsupervised image generation, achieving unprecedented success in generating photorealistic images using adversarial training of generator and discriminator networks. Although conventional use has been towards artistic synthesis or domain adaptation, their potential in solving the ``expensive" problem of data scarcity in scientific imaging—more specifically, in nanomaterial analysis remain largely untapped.
Segmentation of nanoparticles, the operation of detecting and bounding single particles within microscopy images, is indispensable in applications of nanoscience. Segmenting accurately is itself hindered by the rough morphology of the nanoparticles, the aggregates they can form, and stochastic fluctuations of the imaging conditions (e.g., SEM/TEM artefacts). Deep learning-based models, e.g., variants of U-Net \cite{RonnebergerFB15}, have been reported to be good candidates for automating image segmentation but they require access to extensive collection of high fidelity annotated image-mask pairs, which is in great paucity in nanomaterial research.
\\
\\
This paper resolves this bottleneck by reimagining GANs for nanoparticle image generation from segmentation maps. We addressed the challenge of limited annotated data in nanoparticle analysis by employing the cycle-consistent architecture to generate high-fidelity SEM microscopy images directly from segmentation masks. By conditioning the generator on nanoparticle morphology defined by masks, we synthesize diverse and realistic images that mimic stochastic variations in particle size, aggregation patterns, and imaging artifacts. These synthetic images can be used to augment training datasets for downstream segmentation models, overcoming the scarcity of annotated real-world data. This method enabled the generation of high-fidelity, photorealistic nanoparticle microscopy images from masks, significantly expanding limited datasets such as $\mathrm{TiO_2}$-SEM \cite{ruhle2021workflow} through scalable synthesis while improving data quality, thus improving the generalization and accuracy of downstream segmentation models trained on the augmented dataset.
\\
\\
Our significant contributions are as follows:
\begin{itemize}
\item \textbf{Style U-Net implementation:} A novel generator architecture that integrates StyleGAN's photorealistic synthesis with the spatial accuracy of U-Net, improving texture realism while maintaining structural fidelity.
\vspace{2mm}
\item \textbf{Attention-enhanced segmentation:} Addition of self-attention-based mechanisms to prioritize important features within noisy microscopy images.
\vspace{2mm}
\item \textbf{Custom loss functions:} Replacement of conventional cross-entropy (CE) and Dice losses with a combination Focal cross-entropy + Tversky loss to address class imbalance and improve boundary delineation.

\end{itemize}
The rest of this paper is structured as follows: Section \ref{sec:lit} discusses related work in image generation and nanoparticle analysis. Section \ref{sec:method} explains our approach, including architectural advancements and loss functions. Section \ref{sec:data} provides information regarding the dataset used for training our model. Section \ref{sec:result} shows quantitative and qualitative results, followed by conclusion in Section \ref{sec:conclusion}.
\section{Literature Review}
\label{sec:lit}

A critical bottleneck has emerged in the application of deep learning to nanomaterial characterization due to the scarcity of a large-scale annotated dataset. Existing nanoparticle datasets such as $\mathrm{TiO_2}$-SEM \cite{ruhle2021workflow} are insufficient to train decent segment models due to the limited size constraints. Prior attempts to augment smaller datasets with simple image transformations for segmentation by Ronneberger et al. \cite{RonnebergerFB15} showed some significant improvement from previous techniques but these techniques were specifically developed for cell and medical images. The work by Shah et al. \cite{shah2023automated} by incorporating DeepLIFT analysis \cite{shrikumar2017learning} on 2D materials such as graphene pointed out some unique challenges they encountered due to inconsistent sizes and shapes, various contrast ratios, complex background features and shallow field of depth as the model could not simulate the natural variations present in real microscopy images. They concluded that many of these challenges encountered in the classification of 2D materials could be surmounted as the training dataset grows. This indicates that we require more sophisticated systems capable of generating realistic, high-quality nanoparticle images that correspond to their segmentation maps.
\\
Mill et al. \cite{mill2021synthetic} proposed a deep learning workflow for segmentation of complex nanoparticles in high-resolution microscopy images by leveraging a semi-automated synthetic data pipeline based on photorealistic rendering by using softwares such as Blender. Their approach generated an unlimited number of realistic synthetic images with pixel-perfect ground truth labels, enabling the training of convolutional neural networks (CNNs). Similarly, the work by Liang et al. \cite{liang2024segmentation} showed significant improvement in the segmentation of S1 nanoparticles by combining synthetic image-mask pairs with a fraction of real image- mask pairs. They achieved optimal results while training with 15\% authentic and 85\% synthetic data pairs. Although these methods demonstrate the utility of synthetic data for nanomaterial analysis, their dependence on manual input for template generation limits scalability, particularly for diverse or novel nanostructures. This constraint echoes the larger challenges in synthetic data pipelines (Shah et al., 2023 \cite{shah2023automated}), where realism often trades off with generalization.
\\
Unlike manual methods, the Generative Adversarial Network (GAN \cite{goodfellow2014generative}) inherently captures hidden morphological distributions present in real-world microscopy, reducing bias in downstream segmentation. In their initial work, Isola et al. \cite{isola2017image} introduced conditional GANs (cGANs) for paired image-to-image translation and illustrated their application in semantic segmentation and style transfer of images from one form to another retaining significant features. Building on this, Zhu et al. \cite{zhu2017unpaired} introduced CycleGAN, which eliminated the need for paired training data through cycle consistency losses, significantly expanding the applicability of GANs to unpaired image domains. Recently Abzargar et al. \cite{abzargar} followed it up by presenting a custom architecture cGAN-Seg, a special U-Net driven architecture using cycle-consistent losses to both segment cell images and produce high-quality synthetic images from segment maps. This ability to effectively execute bidirectional generation and segmentation is a step forward from conventional cGANs. Although their work proved that cGAN-Seg is effective for cellular image synthesis, we have specifically tailored its generative ability for nanoparticle imaging to attain high fidelity in nanoscale image generation by introducing an attention mechanism and utilizing custom loss functions.
\section{Methodology}
\label{sec:method}
Convolutional neural networks (CNNs), particularly U-Net architectures, have become the norm for segmentation tasks in medical imaging. However, their performance heavily relies on large-scale, high-quality paired datasets (image-ground truth pairs) for robust and accurate training. In real-world scenarios, the acquisition of such datasets is often hindered by costly annotation processes, limited expert availability, and inherent noise in manual labeling, leading to suboptimal generalization during training.
\\
To address this issue, we equipped a cycle consistent generative adversarial network based on cGAN-Seg paper, enhanced with self attention-mechanism in its U-Net segmentation backbone  for realistic image generation and enhanced segmentation accuracy. 

\subsection{Model Construction}
\subsubsection{CycleGAN:}
\begin{figure}
\vspace{-1.7\baselineskip} 
\centering
\includegraphics[width=1\linewidth]{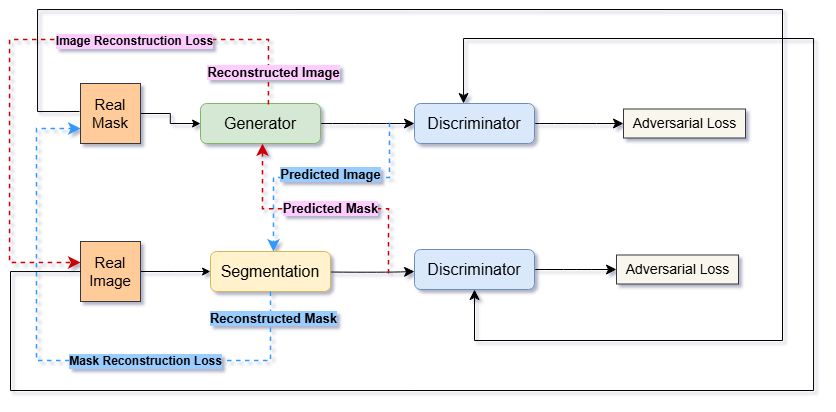}
\caption{Schematic of the proposed generative pipeline. The broken arrows indicate the cycle consistency between the generator and the segmentation network.}
\label{Fig:2}
\end{figure}
Given two image domains $\mathcal{D}_1$ and $\mathcal{D}_2$ CycleGAN is a framework designed for unpaired image-to-image translation, utilizing two interconnected generative adversarial networks (GANs). It consists of two generators—$G:x\rightarrow y$ and  $F:y\rightarrow x$ , which transform the images $x\in \mathcal{D}_1$ and $y\in \mathcal{D}_2$, as well as two discriminators, ${D_x}$ and ${D_y}$, that differentiate between real and generated images in the respective domains. The model is based on two key principles: adversarial training, which ensures that the generated images resemble real samples from the target domain at the optimal training, and cycle-consistency loss, which aim to train the model such that $G(x|x\in \mathcal{D}_1)\in D_2$ and $F(y|y\in \mathcal{D}_2)\in D_1$. As this unconditional mappings are not always achievable independently, we rely on joint training and sample $x\in \mathcal{D}_1$ and $y\in \mathcal{D}_2$ to train the generators such that ${F(G(x))\approx x}$  and ${G(F(y))\approx y}$.\\
The rationale behind employing this architecture in our work is to improve the training data diversity for segmentation models by producing synthetic images and their masks, which are almost impossible to distinguish from actual images. The CycleGAN architecture is distinguished by its two-path training process, involving forward and backward consistency as depicted in Fig-\ref{Fig:2}. In the forward consistency path, the model first transforms a mask image $I_A$ from domain $\mathcal{D}_A$ into nano-particle image $I_B$ in domain $\mathcal{D}_B$ and then tries to transform $I_B$ back to $I_A$ in order to reconstruct the original mask. The backward consistency path also follows the same process but in reverse, beginning with domain $\mathcal{D}_B$, transforming it to domain $\mathcal{D}_A$, and then back to $\mathcal{D}_B$. The bidirectional mapping of this methodology ensures that original images maintain vital characteristics when being translated and reconstructed, eliminating the loss of vital local and global information. 
A segmentation model is then trained on a mix of these artificially produced images and augmented real images.
This hybrid training strategy not only helps to increase the dataset but also upgrades the model's ability to generate realistic images even with fewer samples. By incorporating artificial images, the model sees a broader variety of cases, improving its robustness and generative performance for a wide variety of input masks with significantly different structural contents.
\subsubsection{Style Generation:}
Our proposed architecture has the generator for generating synthetic  $\mathrm{TiO_2}$ images in the form of a 2D-UNET architecture. UNET is largely known for its success with image segmentation due to its unique architecture that  involves a contracting path for obtaining contextual information and a symmetric  expanding path for accurate spatial reconstruction of the image. \\To further boost this design, we have incorporated a style decoding pathway into the decoder part of the UNET architecture, as demonstrated by cGAN-Seg \cite{abzargar}. This is a two- pronged approach that benefits from deriving the best aspects of both architectures— UNET's segmentation prowess and StyleGAN's sophisticated generative  capabilities.
\begin{figure}
\vspace{-0.7\baselineskip} 
\centering
\includegraphics[width=0.9\linewidth]{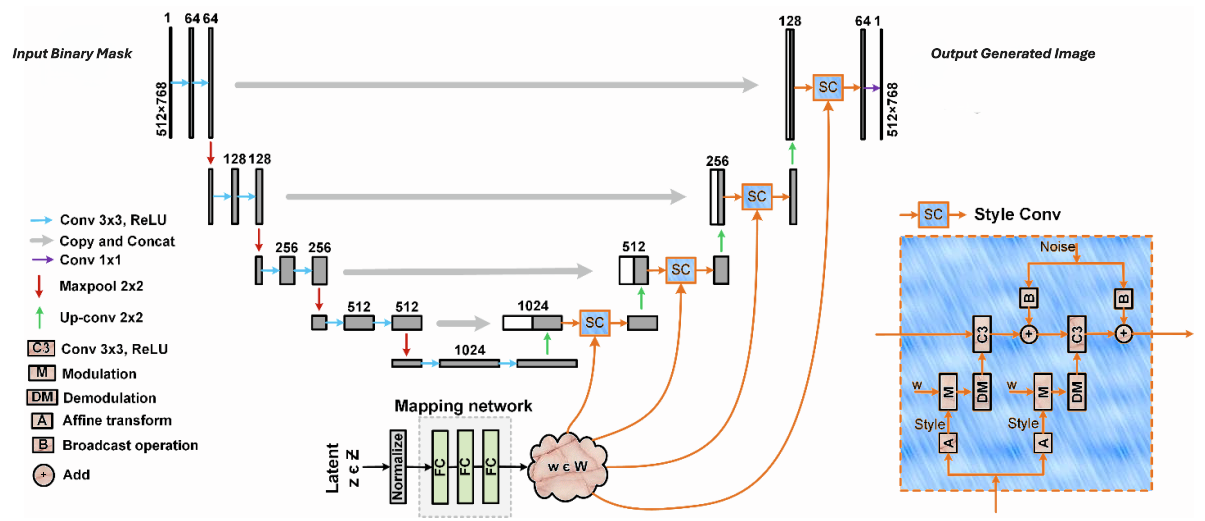}
\caption{Style U-Net Architecture}
\label{Fig:style_unet}
\end{figure}
\\
The style decoding network is implemented to control the stylistic  features of the generated images so that the model can generate more diverse and possibly higher quality synthetic $\mathrm{TiO_2}$ images. 
The core of StyleGAN is a mapping network, a multi-layer perceptron (MLP) that maps a latent vector $z\in \mathcal{Z}$, where $\mathcal{Z}$ is a static noise distribution, to an intermediate style vector $w$. This style vector is then injected into every layer of the synthesis network using adaptive instance normalization (AdaIN) so that there is precise control of stylistic features at different resolutions. In our implementation as depicted in Fig-\ref{Fig:style_unet}, this style vector $w$ is generated by a style-decoding procedure in the decoder of the UNET, in addition to the standard skip connections. AdaIN processing disrupts modulation of the decoder's feature maps by channel-wise alignment (mean and variance) of statistics of those feature maps with learned style parameters. This allows the generator to generate diverse textures and patterns on synthetic $\mathrm{TiO_2}$ images structurally consistent to the contracting pathway of the UNET.\\
In addition, we incorporate noise injection modules at multiple decoder levels to introduce stochastic fine-grained variability. This mimics the inherent variability present in real $\mathrm{TiO_2}$ nanoparticles. Also, our architecture is able to leverage style blending- a technique where independent style vectors: $\ w_i\in W$  are used to control various resolution scales—to separate coarse-grained details from fine-grained textures. Through hierarchical regulation of the UNET's decoder stages, the generator gains the capability to produce images with regulated global morphology (based on the UNET's bottleneck details) and dynamically varying local textures (based on the style path). This blend ensures that the synthetic images have spatial correctness. This blend addresses a key challenge in material science imaging: the structural fidelity vs. generative diversity trade-off. The multi-scale control of the style pathway ensures that synthesized images not only follow the topological constraints learned by the UNET but also have dense, realistic variations in phase distribution and surface roughness.
\\
In our method, we used a modified PatchGAN baseline architecture for our discriminators with an added layer of residual linear attention \cite{abzargar}. 
\begin{figure}
\vspace{-1\baselineskip} 
\centering
\includegraphics[width=0.6\linewidth]{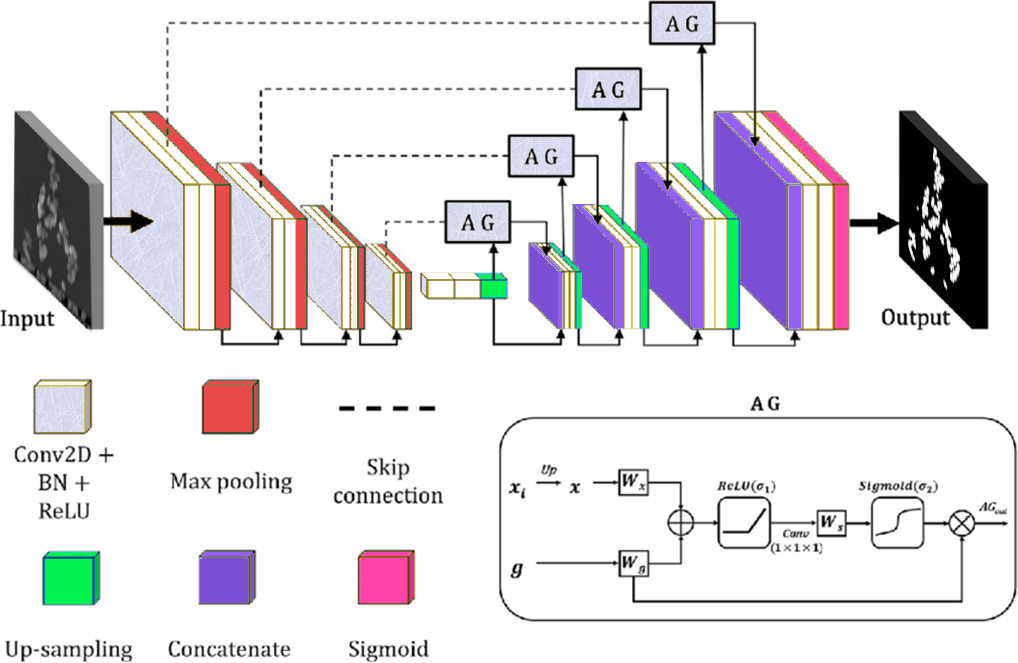}
\caption{Attention U-Net Architecture}
\label{Fig:attn_unet}
\end{figure}
\subsubsection{Segmentation Network:}
For the segmentation task, we incorporated Attention U-Net architecture \cite{RonnebergerFB15} with appropriate routing mechanism as shown in Fig. \ref{Fig:style_unet}. The architecture of our Attention U-Net model is shown in Fig. \ref{Fig:attn_unet}. Through the dynamic weighting of encoder-decoder skip connections, Attention UNet enhances segmentation precision, especially for inconsistently shaped structures.  Segmentation precision is of prime importance as in our overall pipeline, the predicted masks from the $\mathrm{TiO_2}$ images by the Attention UNet network are fed back into the style generator. Hence, it can be inferred that the accuracy of the segmentation network implicitly affects the generation quality of our proposed model.
\subsection{Loss Functions}
\subsubsection{Generation Losses :}
The model employs two complementary loss functions: VGG-based perceptual loss and pixel-wise $\mathcal{L}_1$ loss, both of which are aimed at enhancing various aspects of realistic image generation. The VGG loss employs a pre-trained VGG network for comparing high-level feature representations of synthetic and real images for imposing perceptual consistency in textural details. This encourages the generator to reproduce subtle visual patterns in real microscopy data. The $\mathcal{L}_1$ loss, on the other hand, acts at the pixel level, explicitly minimizing absolute differences between generated and ground-truth images to preserve geometric fidelity in important structural features like nanoparticle size and orientation. 
\subsubsection{Discriminator Loss :}
The Mean Squared Error (MSE) loss is used as the discriminator loss for both the mask discriminator and the image discriminator, providing stable and consistent feedback to the generator. Unlike traditional discriminators, our proposed discriminators use fully supervised losses by computing the squared difference between the discriminator predictions and ground truth target labels. MSE penalizes large errors more harshly than L1, promoting smoother convergence and deterring erratic updates during training. For the image discriminator, this loss encourages pixel-level realism in generated $\mathrm{TiO_2}$ nanoparticles, while the mask discriminator uses MSE to encourage precise segmentation boundaries in synthetic masks. This choice aligns with your goal of generating high-fidelity, structurally accurate outputs, as MSE's geometric alignment emphasis complements the generator's perceptual (VGG) and pixel-wise (L1) losses.
\subsubsection{Segmentation Loss:}
Considering the size and coverage of the $TiO_2$ nanoparticles across images in the dataset, we employed a combination of \textit{Focal Binary Cross-Entropy Loss} and \textit{Tversky Loss}, ensuring a balanced mode of approach towards predicting accurate pixel level classification and tackling the inherent class imbalance present in the dataset. The class-balanced focal cross-entropy loss is defined as:

\begin{equation}
\mathcal{L}_{\text{FCE}} = -\alpha_t (1 - p_t)^\gamma \log(p_t)
\label{eq:fce}
\end{equation}

where $p_t$ is the model's estimated probability for the true class, $\alpha_t \in [0,1]$ is the class balancing weight and $\gamma \geq 0$ is the focusing parameter that down-weights easy examples. The boundary-aware Focal Tversky loss is computed as:
\begin{equation}
\mathcal{L}_{\text{TV}} = (1 - \text{TI})
\label{eq:ft}
\end{equation}
where the Tversky Index (TI) measures segmentation overlap as $\text{TI} = \frac{TP}{TP + \alpha FP + \beta FN}$, with $TP$, $FP$, $FN$ denoting true positives, false positives, false negatives, and  $\alpha$ \& $\beta$ controlling the false positive/negative trade-off.\\
The total segmentation loss is defined as:

\begin{equation}
\mathcal{L}_{\text{Total}} = \lambda_1 \mathcal{L}_{\text{FCE}} + \lambda_2 \mathcal{L}_{\text{TV}}
\label{eq:total_loss}
\end{equation}
where $\lambda_1$ and $\lambda_2$ are relative weighting factors influencing the contribution of each of these respective components towards the entire loss function. In order to achieve optimal results we assigned equal weightage with $\lambda_1 = \lambda_2 = 1$.
\section{Dataset and Data Pre-processing}
\label{sec:data}
\subsection{Dataset}
The data used in this research is obtained from the GitHub repository(\url{https://github.com/BAMresearch/automatic-sem-image-segmentation}). The dataset includes electron microscopy (EM) images of $\mathrm{TiO_2}$ particle and their respective segmentation masks defining the nanoparticle boundaries.The repository is organized into sub-folders with raw scanning electron microscopy (SEM) and transmission scanning electron microscopy (TSEM) images and their respective manually labelled segmentation and classification masks. 

%
\subsection{Data Pre-processing}
To facilitate consistency in input data for the neural network, the SEM images were pre-processed using methods like scaling and normalization. Resizing scales the images to a constant size, and normalization rescales pixel values to match the needs of the network, thus maximizing computational effectiveness. These normalization processes make the model more resilient and capable in dealing with variations in nanoparticle morphology usually found in SEM images. Augmentation techniques like horizontal and vertical flipping, CLAHE, random crop etc. were employed to ensure that the model stays robust to variations and identifies the important patterns in data distributions that might vary from the training dataset in real world. In order to ensure efficient training and reliable testing of the model's capabilities, data was divided into 3 sets, namely training, validation and test sets. 70\% of the entire dataset was utilised for training, while 20\% and 10\% were allocated for testing and validation respectively.
\subsection{Training Configuration}
The model was trained for 700 epochs using the Adam optimizer with a learning rate of 0.0001 and a batch size of 2. These parameters were selected through empirical validation to balance convergence stability and computational efficiency for generation tasks.
%
%
All the experiments were conducted on an Intel Xeon Silver 4216 CPU (2.10GHz), NVIDIA Quadro RTX 6000 GPU (24GB VRAM), and 64GB DDR4 RAM (2400MHz), with storage on Micron 1300 SATA SSDs (512GB).
\section{Experimental Results}
\label{sec:result}
\subsection{Evaluation Metrics}
\subsubsection{FID Score:}
The Fréchet Inception Distance (FID) score is a typical evaluation of generated image quality based on how much the distribution of real images differ from that of generated images, initially introduced by Heusel et al.\cite{heusel2017gans}. It estimates the Fréchet distance between feature representations derived from a pre-trained Inception network, considering both mean and covariance shifts between real and generated images. Lower FID scores indicate higher visual fidelity and diversity in the produced images. Hence, we employ FID on the model to measure the quality of images produced by us quantitatively and compare different model configurations based on this score.
\subsubsection{SSIM Score:}
The Structural Similarity Index Measure (SSIM) is a metric widely used for assessing the perceptual quality of images by comparing the structural information between a reference image and a generated image. Introduced by Wang et al. \cite{wang2004image}, SSIM evaluates luminance, contrast, and structure independently and combines them into a single score. Higher SSIM values (closer to 1) indicate better perceptual similarity to the reference image. Hence, we employ SSIM to quantitatively measure the similarity between generated images and their real counterpart images.

\subsection{Results}
\begin{figure}[htbp]
    \centering    
    \begin{minipage}[t]{0.18\textwidth}
        \centering
        
        \text{}\\
        \text{}\\
        \vspace{1.5mm}
        
        \includegraphics[width=\linewidth, height=3cm]{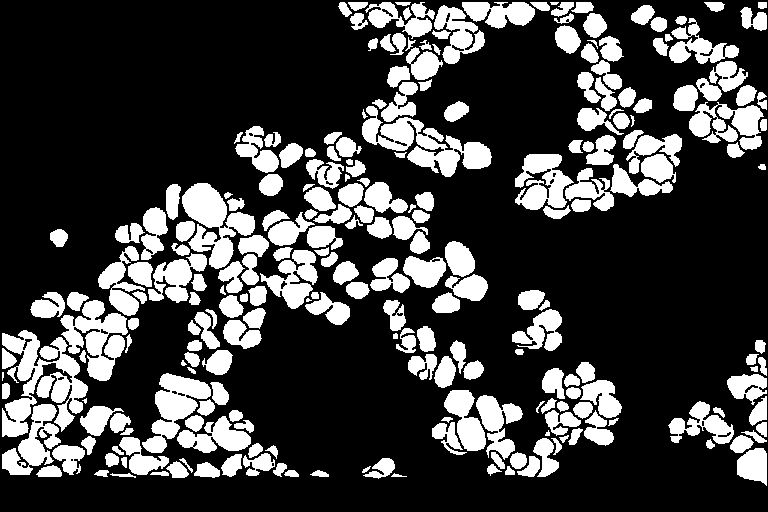}

        \vspace{1.4mm} 
        
        \includegraphics[width=\linewidth, height=3cm]{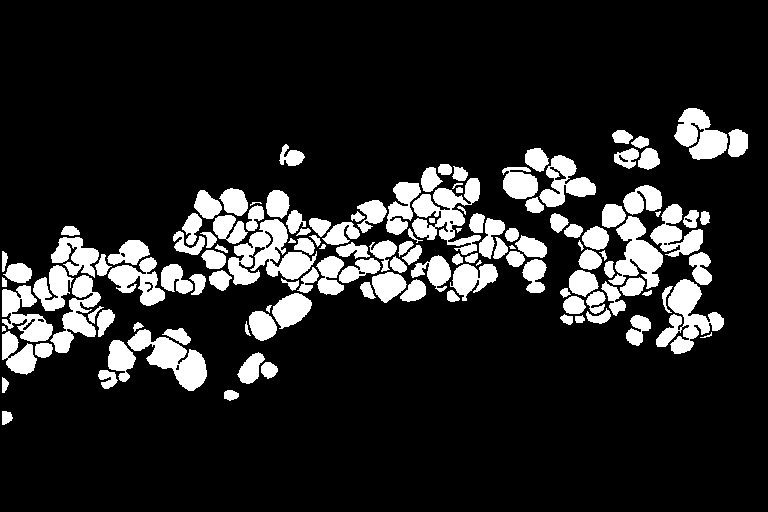}
        {Real Mask}\\
        \label{fig:typeA}
    \end{minipage}\hspace{0.005\textwidth}
    \begin{minipage}[t]{0.18\textwidth}
        \centering
        
        \text{}\\
        \text{}\\
        \vspace{1.5mm}
        
        \includegraphics[width=\linewidth, height=3cm]{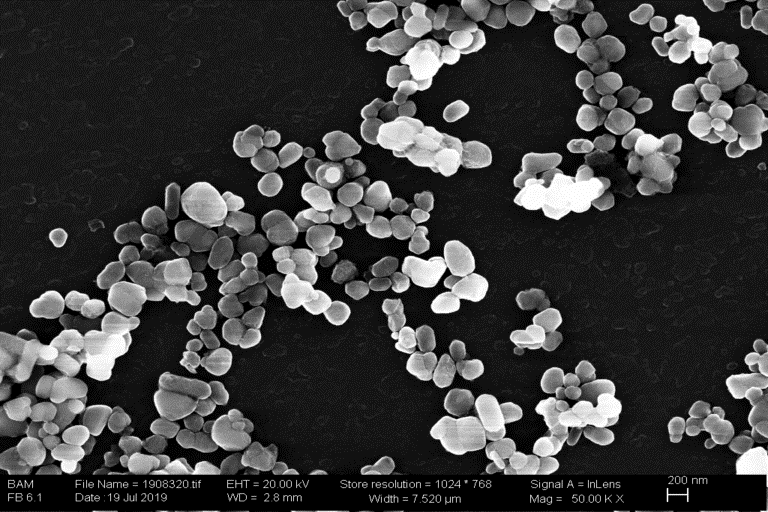}
        
        \vspace{1.4mm}
        
        \includegraphics[width=\linewidth, height=3cm]{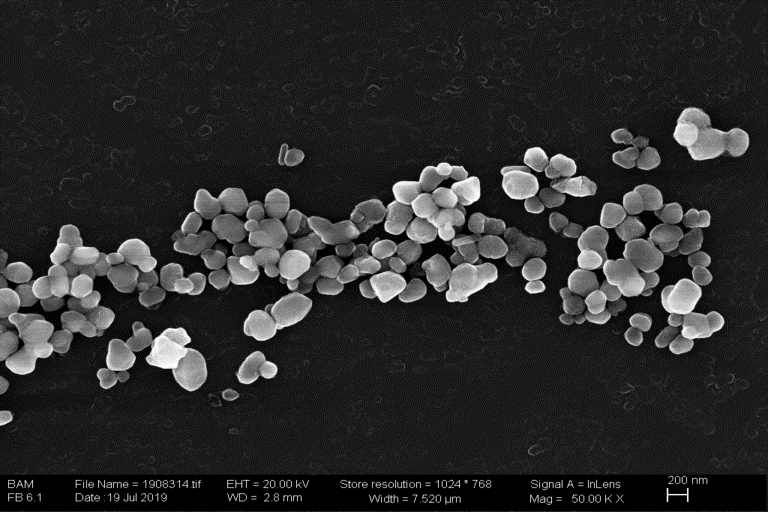}
        {Real Image}\\
        \label{fig:typeB}
    \end{minipage}\hspace{0.005\textwidth}
    \begin{minipage}[t]{0.18\textwidth}
        \centering
        
        \text{FID: 69.90}\\
        \text{SSIM: 0.189}\\
        \vspace{1.5mm}
        
        \includegraphics[width=\linewidth, height=3cm]{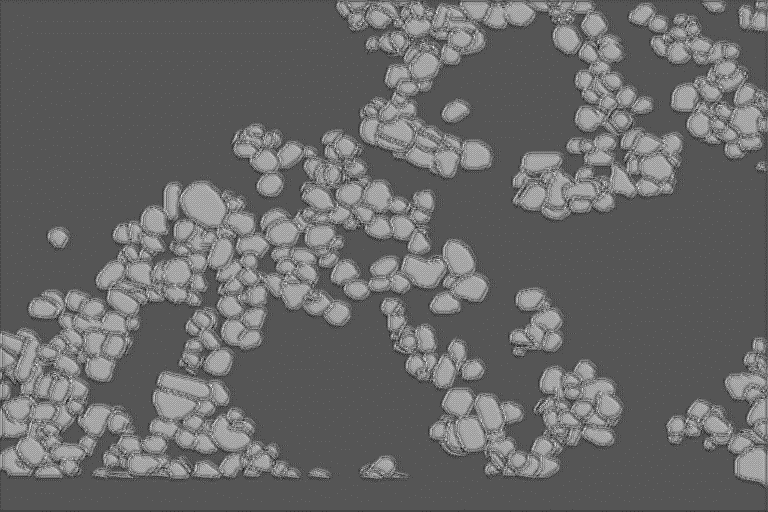}
        
        \vspace{1.4mm}
        
        \includegraphics[width=\linewidth, height=3cm]{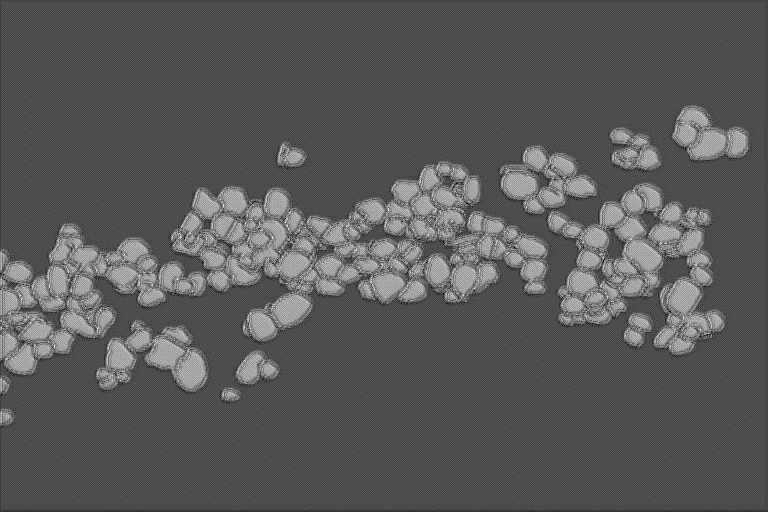}
        {GAN}\\
        
        \label{fig:typeC}
    \end{minipage}\hspace{0.005\textwidth}
    \begin{minipage}[t]{0.18\textwidth}
        \centering
        
        \text{FID: 52.01}\\
        \text{SSIM: 0.369}\\
        \vspace{1.5mm}
        
        \includegraphics[width=\linewidth, height=3cm]{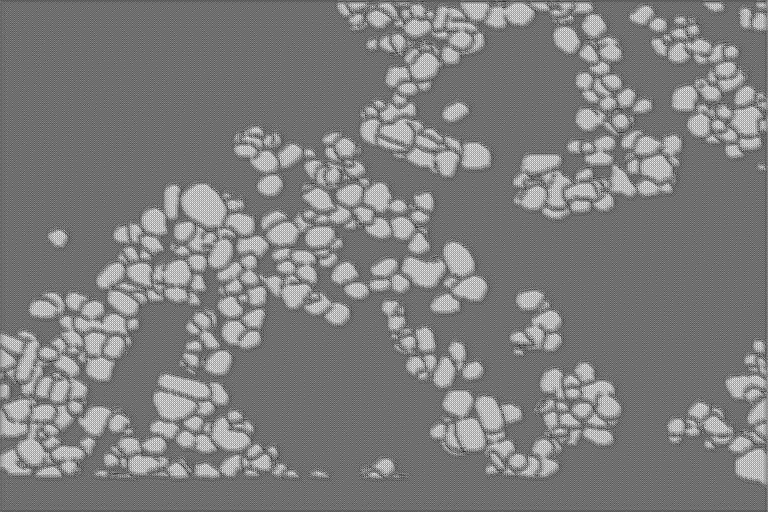}
        
        \vspace{1.4mm}
        
        \includegraphics[width=\linewidth, height=3cm]{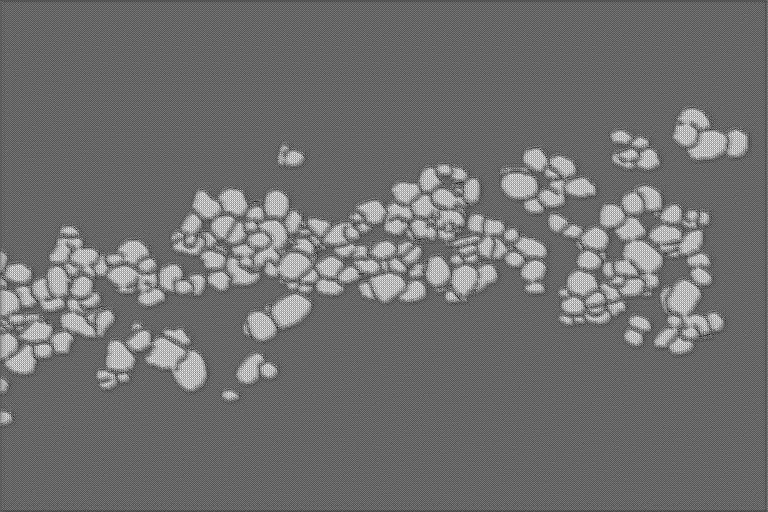}
        {Cycle GAN}\\
        
        \label{fig:typeD}
    \end{minipage}\hspace{0.005\textwidth}
    \begin{minipage}[t]{0.18\textwidth}
        \centering
        
        \textbf{FID: 17.65}\\
        \textbf{SSIM: 0.546}\\
        \vspace{1.5mm}
        
        \includegraphics[width=\linewidth, height=3cm]{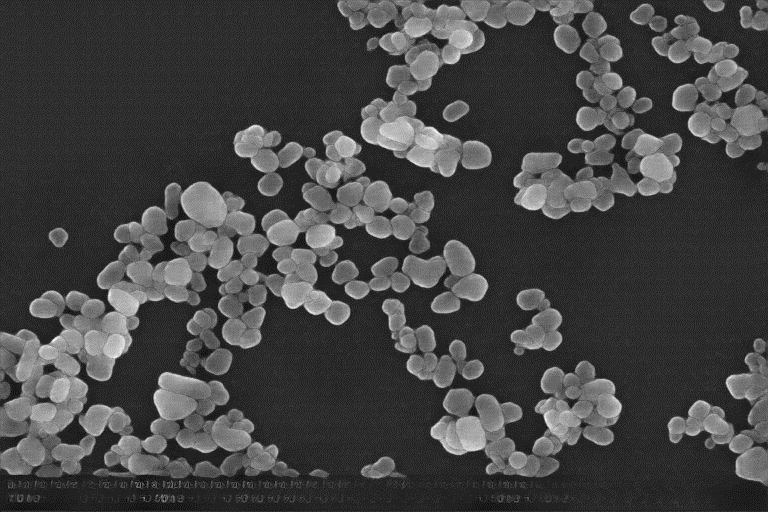}

        \vspace{1.4mm}
        
        \includegraphics[width=\linewidth, height=3cm]{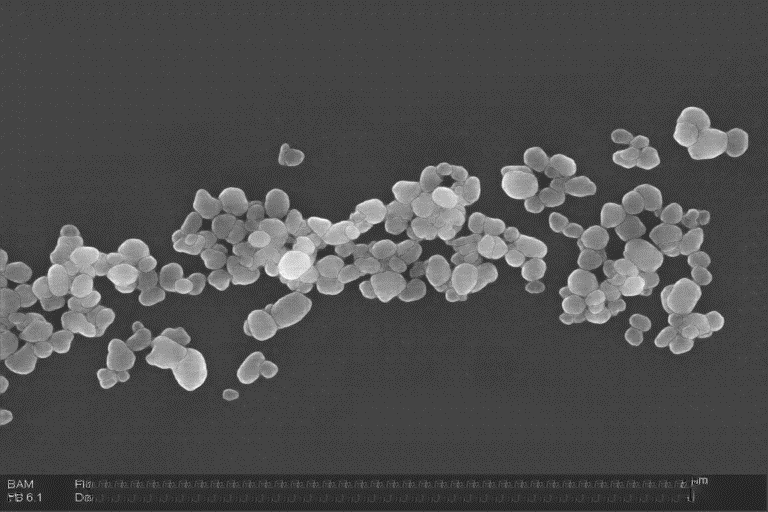}
        \textbf{F-ANcGAN}\\
        
        \label{fig:typeE}
    \end{minipage}   
    \caption{Comparison of our proposed model against standard models}
    \label{fig:Comp_mod}
\end{figure}

\begin{figure}
\centering
\begin{subfigure}{0.45\textwidth}
\centering
\includegraphics[width=0.97\textwidth]{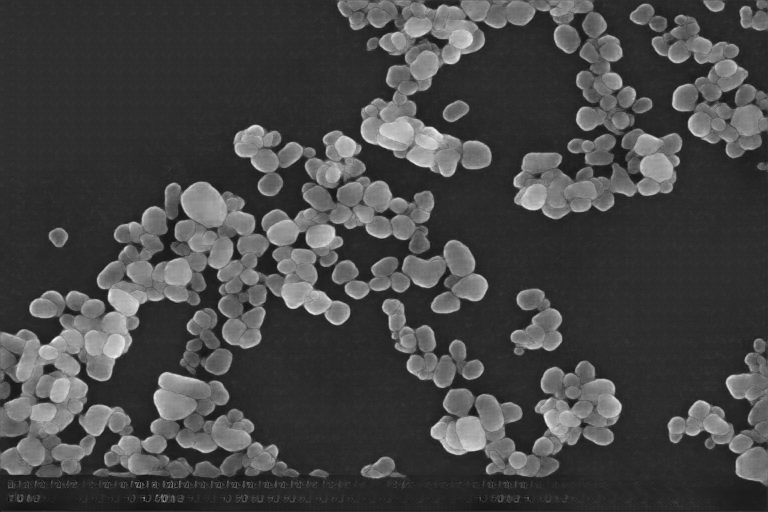}
\caption{Raw generated image \textbf{FID- 17.65}}
\end{subfigure}
\begin{subfigure}{0.45\textwidth}
\centering
\includegraphics[width=0.97\textwidth]{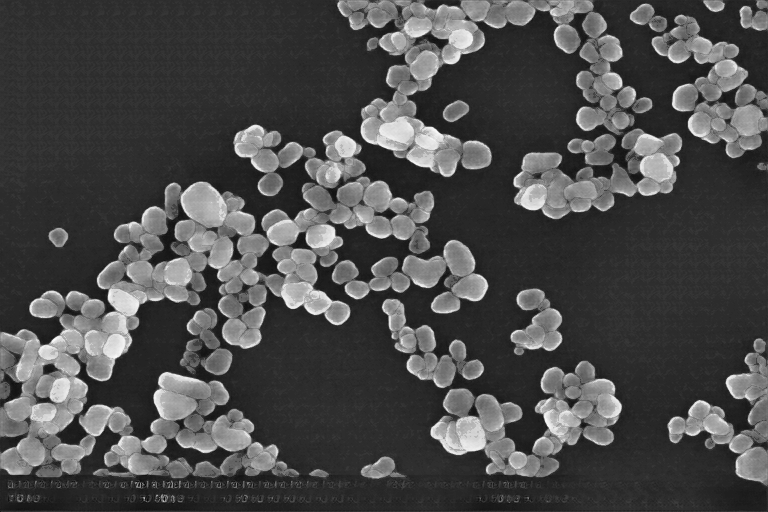}
\caption{Post-processed generated image \textbf{FID- 10.39}}
\end{subfigure}
\caption{Raw generated image vs Post-processed generated image from $\mathrm{TiO_2}$ dataset}
\label{fig:post-process}
\end{figure}

\begin{figure}[htbp]
    \centering    
    \begin{minipage}[t]{0.3\textwidth}
    \textbf{}
        \centering\\
        \vspace{1mm}
        \includegraphics[width=\linewidth, height=3cm]
        {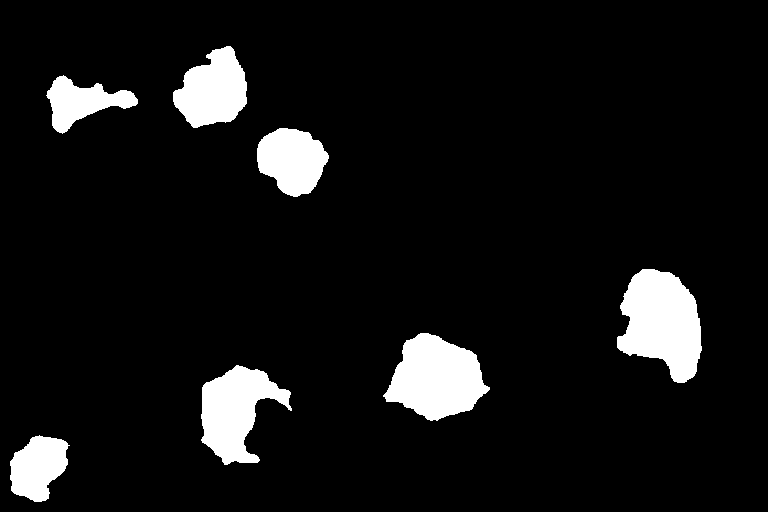}

        \vspace{1.4mm}
        
        \includegraphics[width=\linewidth, height=3cm]{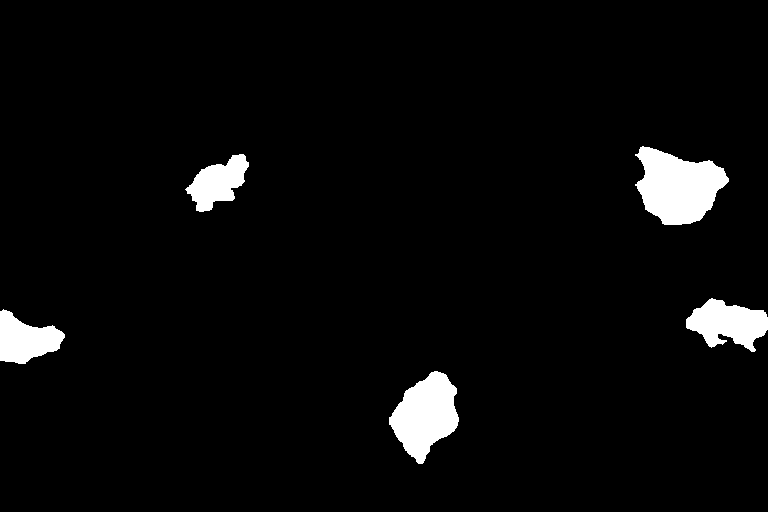}
        {Real Mask}
        \label{fig:typeA}
    \end{minipage}\hspace{0.005\textwidth}
    \begin{minipage}[t]{0.3\textwidth}
        \centering
        \textbf{}\\
        \vspace{1mm}
        \includegraphics[width=\linewidth, height=3cm]{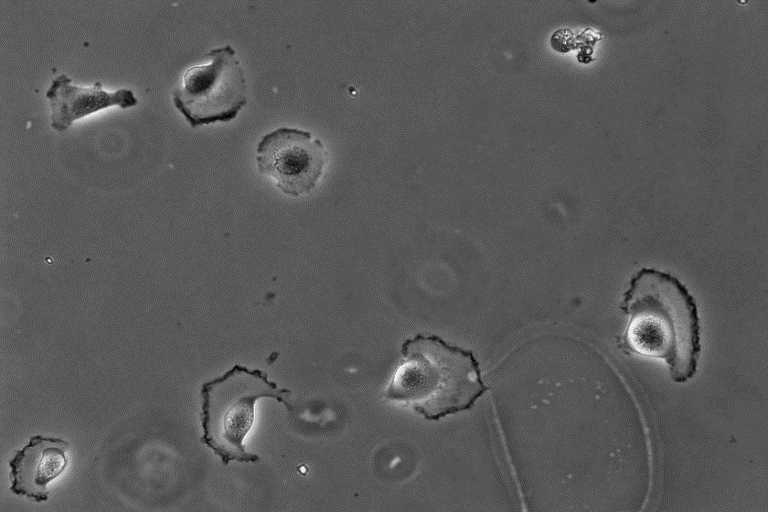}

        \vspace{1.4mm}
        
        \includegraphics[width=\linewidth, height=3cm]{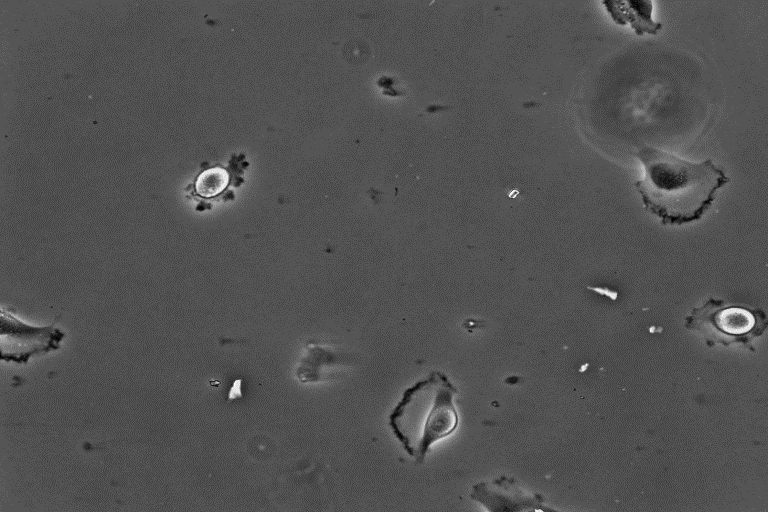}
        {Real Image}
        \label{fig:typeB}
    \end{minipage}\hspace{0.005\textwidth}
    \begin{minipage}[t]{0.3\textwidth}
        \centering
        \textbf{FID: 34.69}\\
        \vspace{1mm}
        \includegraphics[width=\linewidth, height=3cm]{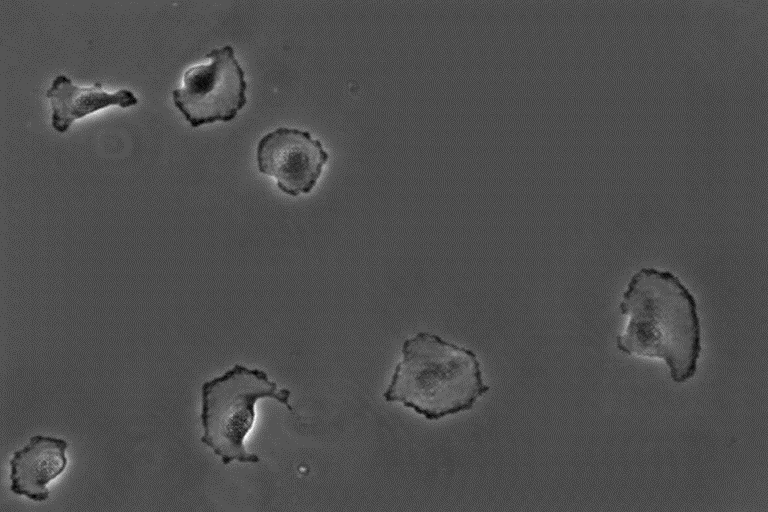}

        \vspace{1.4mm}
        
        \includegraphics[width=\linewidth, height=3cm]{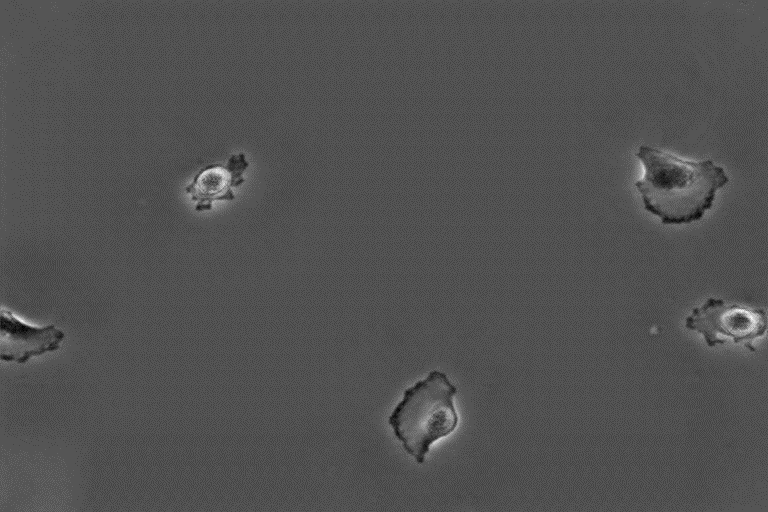}
        {\textbf{F-ANcGAN} Generated Image}\\
        \label{fig:typeC}
    \end{minipage}\hspace{0.005\textwidth} 
    \caption{Generation of PHC-C2DH-U373 dataset images containing human glioblastoma cells}
    \label{fig:gen_mod}
\end{figure}

So, we sought to evaluate our model on the test dataset that we had made from the original dataset, which was completely unseen by the model during its course of training and validation.  The synthetic images generated by the original masks of the test set had an average Frechet Inception Distance (FID) score of 17.65 and an SSIM score of 0.546, indicating a high degree of similarity between the real and synthetic $\mathrm{TiO_2}$ images as highlighted in Fig-\ref{fig:Comp_mod}. Notably with some minor targeted post-processing adjustments of brightness, exposure and shadow/highlight balance, the F-ANcGAN's output of $\mathrm{TiO_2}$ nanaparticle (Fig-\ref{fig:post-process} ) achieved a 41\% FID improvement (17.65 → 10.39).
\\
\begin{table}[h]
\centering
\vspace{-1\baselineskip} 
\caption{FID comparisons between models}
\vspace{1mm}

\begin{tabular}{|p{7cm}|p{3cm}|} 
\hline\centering
\textbf{METHODOLOGY} & \centering\textbf{FID SCORE} \tabularnewline\hline\centering
Generative Adversarial Network (GAN) & \centering 69.90 \tabularnewline \hline\centering
Cycle Consistent Generative Adversarial Network (CYCLEGAN) & \centering52.01 \tabularnewline \hline\centering
\textbf{Proposed Model: F-ANcGAN} & \centering\textbf{17.65} \tabularnewline \hline
\end{tabular}
\label{tab:FID_Comparisons}
\end{table}\\
We conducted a comparative study using standard generative models, namely GAN and CycleGAN by training them on the $\mathrm{TiO_2}$ dataset in order to compare their generation capabilities with our model as depicted in Table-\ref{tab:FID_Comparisons}. The results clearly highlight our model's generation capacity as our FID score markedly surpasses the GAN architecture(69.90) and the more sophisticated CycleGAN architecture(52.01).\\
The enhancement in metrics by our proposed model, we infer, is due to the attention mechanism that is incorporated in the 2D Unet Segmentation network, whose enhanced ability to focus on relevant regions of the image implicitly drives the Style-Unet generator's ability to generate synthetic images that are closer in style and texture to the real images.
Our modified architecture exhibits increased skill in working with varying ranges of depth and exposure as present in our training dataset, positioning itself as a reliable and efficient solution for highly realistic generation tasks in nanomaterial science.\\
\\
To demonstrate the generalization of our proposed approach, we trained our model on a medical dataset \textbf{PhC-C2DH-U373} (\url{https://celltrackingchallenge.net/2d-datasets/}) which comprises of phase-contrast microscopy images of U-373 MG human glioblastoma cells cultured on a polyacrylamide substrate. These glioblastoma cells were completely diverse in style, shape and texture from the $\mathrm{TiO_2}$  nanoparticle dataset our model was solely trained upon. Despite the varied data distribution of this dataset, our model still generated synthetic cell images similar to the real domain images, accomplishing a relatively low FID score (Fig-\ref{fig:gen_mod}).\\

\begin{figure}[htbp]
    \centering  
    \begin{minipage}[t]{0.3\textwidth}
        \centering
        \includegraphics[width=\linewidth, height=3cm]{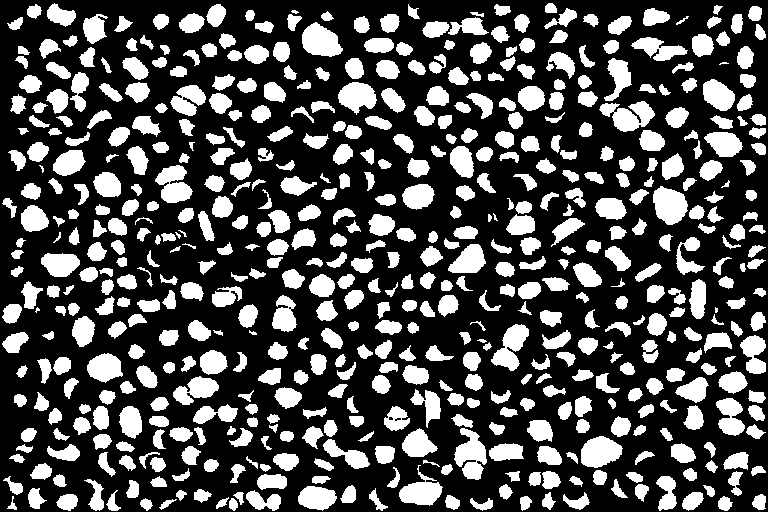}

        \vspace{1.4mm}
        
        \includegraphics[width=\linewidth, height=3cm]{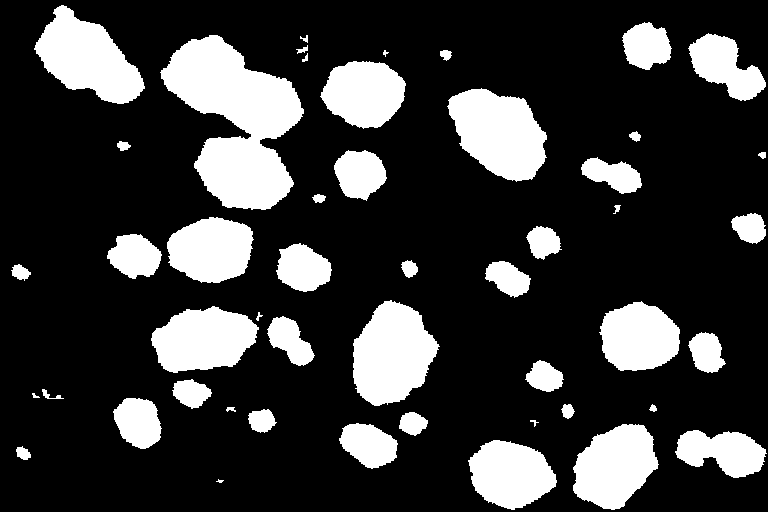}
        {High Density Mask}
        \label{fig:typeA}
    \end{minipage}\hspace{0.005\textwidth}
    \begin{minipage}[t]{0.3\textwidth}
        \centering
        \includegraphics[width=\linewidth, height=3cm]{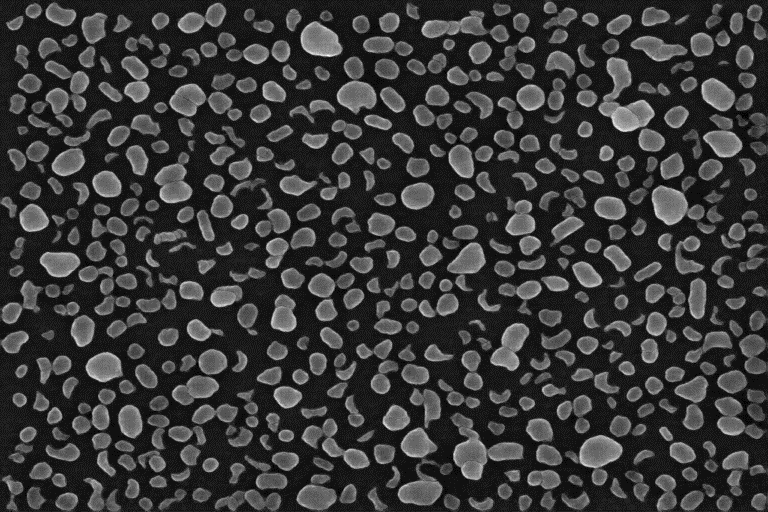}

        \vspace{1.4mm}
        
        \includegraphics[width=\linewidth, height=3cm]{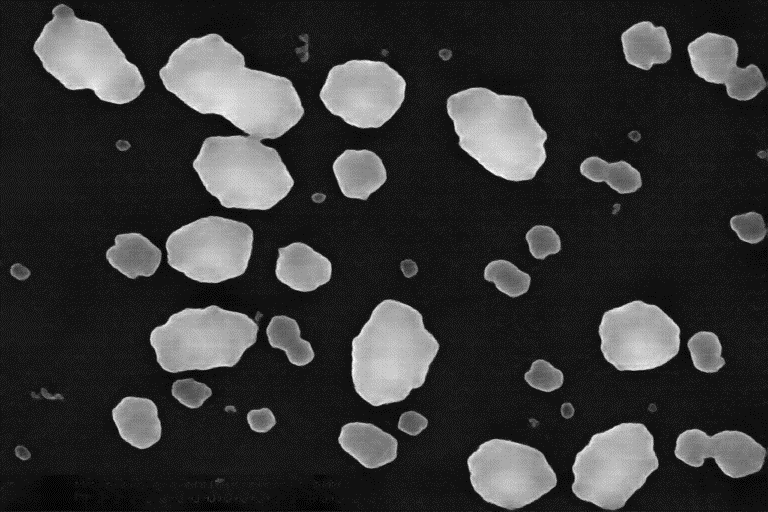}
        {Generated Image}
        \label{fig:typeB}
    \end{minipage}\hspace{0.005\textwidth} 
    \caption{Synthetic high density nanoparticle-like masks and their counterpart generated images}
    \label{fig:gen_mod_2}
\end{figure}

The $\mathrm{TiO_2}$ nanoparticles dataset consisted of low- and mid- density clusters of nanoparticle images. We were interested to know if our generator could apply its knowledge by generating new synthetic images comprising   high-density clusters of nanoparticle-like images that were not part of the training distribution. To achieve this, we generated synthetic high density nanoparticle masks and fed them as input to our generator in its testing phase. As depicted in Fig-\ref{fig:gen_mod_2}. , our model was able to extrapolate its knowledge from low and mid-density nanoparticle images and accurately generate synthetic high-density nanoparticle images. Such a learning capacity enables us to generate synthetic nanoparticle images which adhere to a wide range of real-world scenarios. The ability of our model to extrapolate learned knowledge to unseen scenarios makes it a powerful tool for generalization,  potentially enabling the development of generation and segmentation models for a variety of image datasets cutting across modalities.

\subsection{Ablation Studies}
To ensure the validity of our choices made in loss functions, we performed an ablation study comparing the effects of various segmentation loss functions on model generation performance. Our setup that uses Focal Cross-Entropy (CE) and Tversky loss ($\alpha$ =0.4, $\beta$=0.6) produced the best results with an FID of 17.65 and an SSIM of 0.546, proving that penalizing false positives moderately ($\alpha$=0.4) with a greater emphasis on false negatives ($\beta$=0.6) works best to retain nanoparticle morphology while ensuring image realism. The Focal Tversky variant ($\gamma$=0.75) with adjusted weights ($\alpha$=0.3, $\beta$=0.7) demonstrated better structural similarity (SSIM=0.654) but underperformed in terms of image quality (FID=21.92), indicating that while focal weighting of Tversky Loss assists in enhancing structural details, it could potentially introduce artifacts. The Dice loss baseline was the weakest of the three (FID=22.70, SSIM=0.432), affirming its struggles in addressing nanomaterial imaging challenges where accurate boundary detection is essential. These findings as depicted in Table-\ref{tab:ablation} show that meticulous tuning of the Tversky loss parameters along with Focal Cross-Entropy loss yields superior generation of synthetic images both qualitatively and quantitatively for nanoparticle analysis.
\begin{table}[h]
\centering
\caption{Ablation Study}
\vspace{1mm}
\begin{tabular}{|p{7cm}|p{3cm}|p{3cm}|} 
\hline\centering
\textbf{Loss Configuration} & \centering\textbf{FID SCORE} & \centering\textbf{SSIM SCORE} \tabularnewline\hline\centering
Focal CE + Dice & \centering 22.70 & \centering 0.432 \tabularnewline \hline\centering
CE + Focal TV($\alpha$=0.3, $\beta$=0.7, $\gamma$=0.75) & \centering 21.92 & \centering\textbf{0.654} \tabularnewline \hline\centering
\textbf{Focal CE + TV($\alpha$=0.4, $\beta$=0.6)} & \centering \textbf{17.65} & \centering 0.546 \tabularnewline \hline
\end{tabular}
\label{tab:ablation}
\end{table}
\subsection{Human Evaluation}
\begin{figure}
\centering
\begin{subfigure}{0.45\textwidth}
\includegraphics[width=\textwidth]{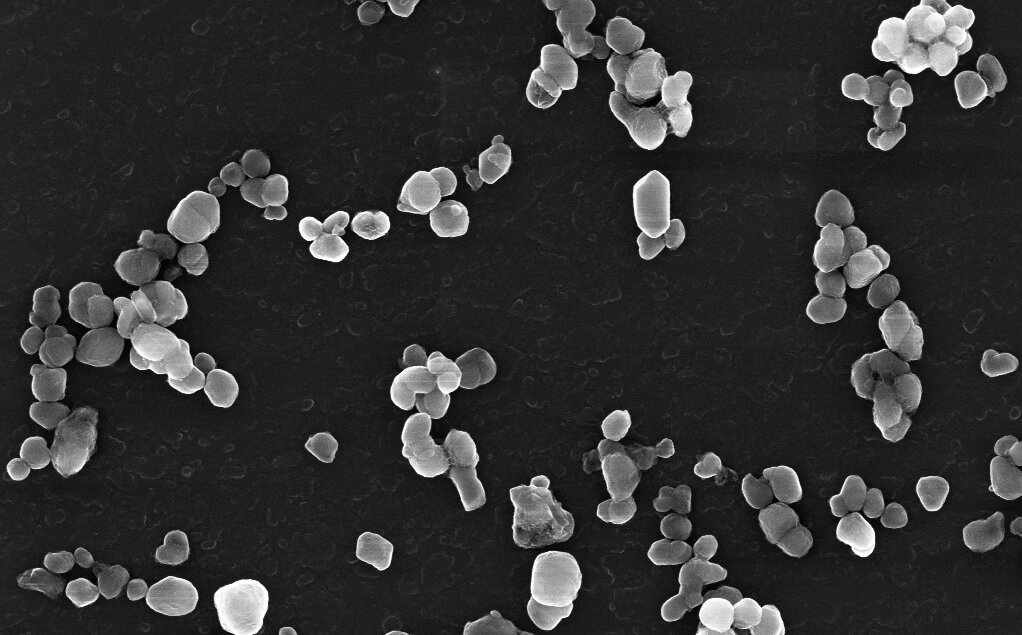}
\caption{Real Image}
\end{subfigure}
\hspace{1.4mm}
\begin{subfigure}{0.45\textwidth}
\includegraphics[width=\textwidth]{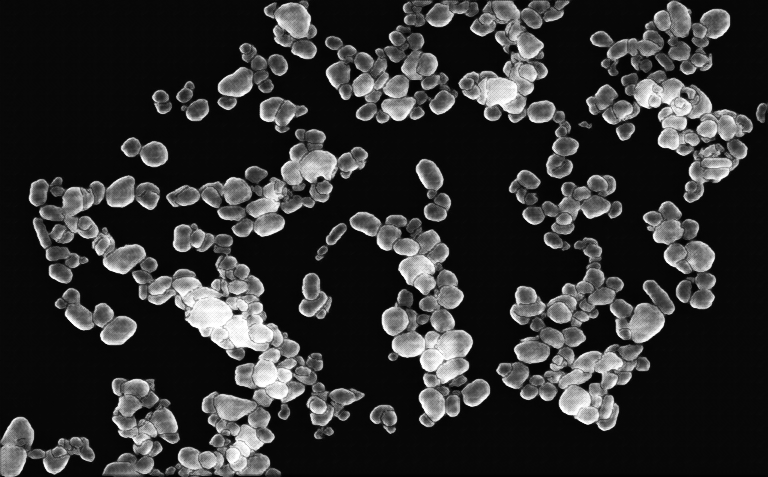}
\caption{Generated Image}
\end{subfigure}
\centering
\caption{Real and generated image samples from human evaluation test set}
(To be noted that both these images are independent of each other and so are the subsequent images of the set)
\label{fig:Human_eval_test}
\end{figure}
To assess the perceptual quality of the synthetic images generated by our model, we conducted a human evaluation study. A set of 16 images— comprising 8 real images randomly sampled from our original $\mathrm{TiO_2}$ dataset and 8 randomly sampled synthetic images produced by our proposed model—was presented to human evaluators in a randomized order. A real and generated image used in our human evaluation set is shown as reference in Fig-\ref{fig:Human_eval_test} and a couple more samples of real and generated images are shown in Fig S1. Prior to the evaluation, participants were shown 10 pairs of real images along with their corresponding ground truth masks to help them become familiar with the visual characteristics of the dataset. Importantly, the evaluators were not informed about whether any image in the evaluation set was real or synthetic and the evaluation process was not time-bound in any manner. Each image received multiple independent ratings, resulting in a total of 640 responses. This setup allowed us to investigate how convincingly the synthetic images could emulate real ones from the perspective of human perception.\\
\begin{figure}
\vspace{0.5\baselineskip} 
\centering
\includegraphics[width=0.45\linewidth]{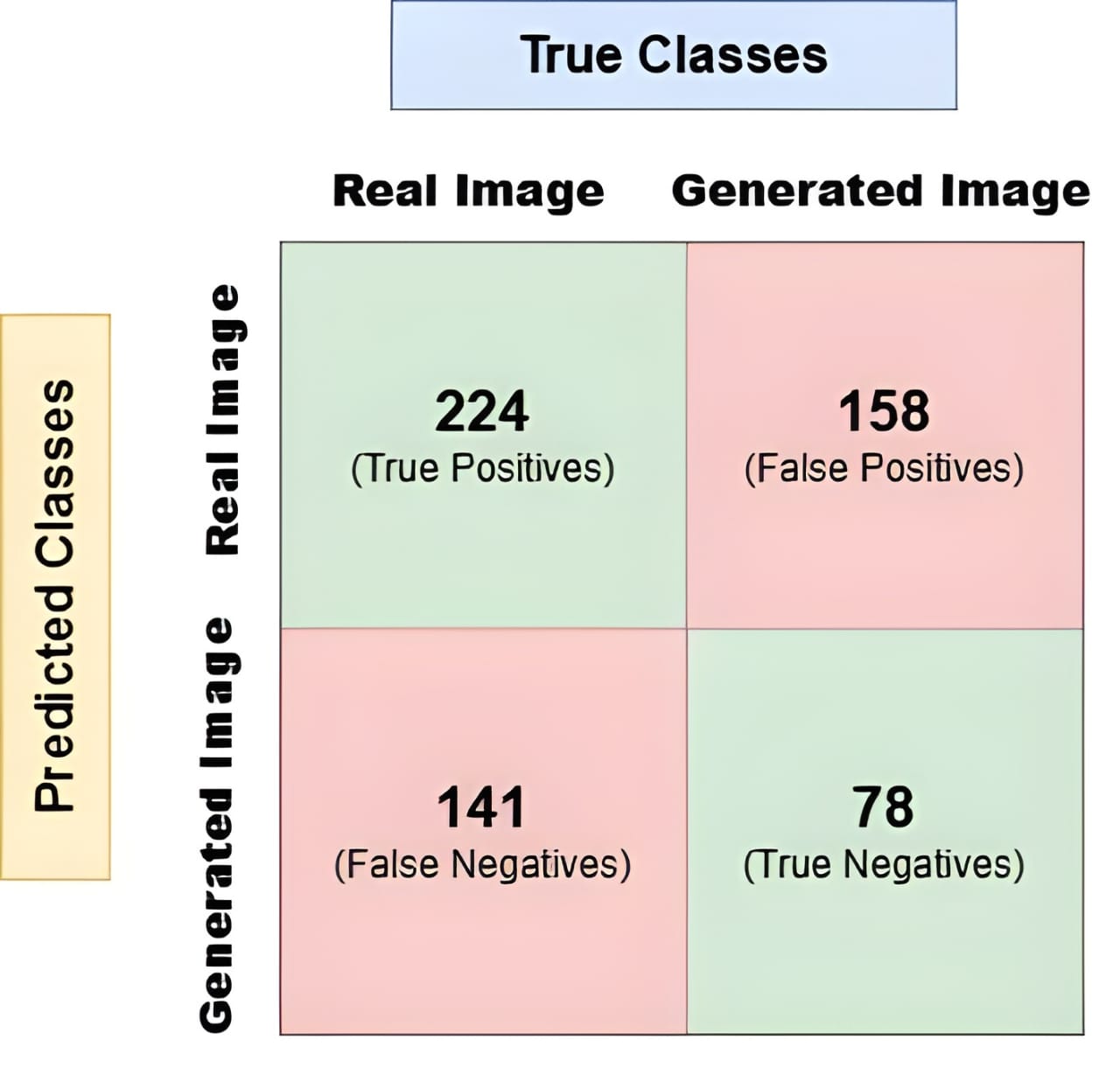}
\caption{Confusion Matrix of Human Evaluation Results}
\label{Fig:confusion_matrix}
\end{figure}
Each image in the evaluation set was presented with three response options: \textit{\textbf{Real}}, \textit{\textbf{Fake}}, and \textit{\textbf{Not Sure}}, allowing evaluators to assess the image and indicate whether they believed it was real, synthetic or they were uncertain of its nature. Out of the 640 total responses, 39 were marked as \textit{\textbf{Not Sure}}, leaving 601 confident responses for further analysis. To better understand the evaluators' ability to distinguish between real and synthetic images, we constructed a confusion matrix(Fig-\ref{Fig:confusion_matrix}) based on these confident responses.\\

\section{Conclusion}
\label{sec:conclusion}
This work presents a more advanced cGAN-Seg architecture named F-ANcGAN, developed to achieve realistic generation of high-quality microscopy images directly from segmentation maps. By incorporating self-attention mechanisms to model long-range spatial relationships and by utilizing focal loss functions to tackle class imbalance, our method achieves a high degree of generative realism , as demonstrated by an FID score of 17.65 and an SSIM score of 0.546 on $\mathrm{TiO_2}$ nanoparticle dataset (Fig. \ref{fig:Comp_mod}). In particular, the model also exhibits good generalization abilities, effectively synthesizing realistic glioblastoma cell images (Fig. \ref{fig:gen_mod}), which also attests to its versatility in handling various biomedical contexts. A main limitation comes with reconstructing delicate background details, since segmentation masks inherently do not have explicit background data, resulting in the generator's focus on foreground accuracy. Nevertheless, our method considerably improves segmentation-to-image translation for nanoscale and biomedical contexts, especially by maintaining vital foreground structures. Additionally, our approach significantly reduces the reliance on large annotated datasets, enabling more reliable downstream segmentation tasks. Our work closes the gap between synthetic and real data domains, providing an efficient and scalable approach for nanomaterial research where data availability is a relevant constraint.

%

\bibliography{reference}
\end{document}